\documentclass[10pt,twocolumn,letterpaper]{article}

\usepackage{cvpr}              % To produce the CAMERA-READY version
\usepackage{multirow}
\definecolor{cvprblue}{rgb}{0.21,0.49,0.74}
\usepackage[pagebackref,breaklinks,colorlinks,allcolors=cvprblue]{hyperref}
\usepackage{arydshln}
\def\paperID{12} % *** Enter the Paper ID here
\def\confName{CVPR}
\def\confYear{2025}

\title{Are Vision-Language Models Ready for Dietary Assessment? \\Exploring the Next Frontier in AI-Powered Food Image Recognition}

\author{Sergio Romero-Tapiador$^{1}$\and Ruben Tolosana$^{1}$\and Blanca Lacruz-Pleguezuelos$^{2}$\and  Laura Judith Marcos-Zambrano$^{2}$\and Guadalupe X. Bazán$^{2}$\and Isabel Espinosa-Salinas$^{2}$\and Julian Fierrez$^{1}$ \and Javier Ortega-Garcia$^{1}$ \and Enrique Carrillo de Santa Pau$^{2}$ \and Aythami Morales$^{1}$ \and {\normalsize $^1$Biometrics and Data Pattern Analytics Lab, Universidad Autonoma de Madrid, Madrid, Spain} \and
{\normalsize $^2$IMDEA Food, CEI UAM+CSIC, Madrid, Spain}
}

\begin{document}
\maketitle
\begin{abstract}
Automatic dietary assessment based on food images remains a challenge, requiring precise food detection, segmentation, and classification. Vision-Language Models (VLMs) offer new possibilities by integrating visual and textual reasoning. In this study, we evaluate six state-of-the-art VLMs (ChatGPT, Gemini, Claude, Moondream, DeepSeek, and LLaVA), analyzing their capabilities in food recognition at different levels. For the experimental framework, we introduce the FoodNExTDB, a unique food image database that contains  9,263 expert-labeled images across 10 categories (e.g., ``protein source"), 62 subcategories (e.g., ``poultry"), and 9 cooking styles (e.g., ``grilled"). In total, FoodNExTDB includes 50k nutritional labels generated by seven experts who manually annotated all images in the database.  Also, we propose a novel evaluation metric, Expert-Weighted Recall (EWR), that accounts for the inter-annotator variability. Results show that closed-source models outperform open-source ones, achieving over 90\% EWR in recognizing food products in images containing a single product. Despite their potential, current VLMs face challenges in fine-grained food recognition, particularly in distinguishing subtle differences in cooking styles and visually similar food items, which limits their reliability for automatic dietary assessment. The FoodNExTDB database is publicly available at \href{https://github.com/AI4Food/FoodNExtDB}{https://github.com/AI4Food/FoodNExtDB}.
\end{abstract}
    
\section{Introduction}
\label{sec:intro}

Food is a fundamental source of energy for human life and plays a critical role in preventing chronic diseases. Maintaining a healthy diet has become increasingly challenging due to multiple factors, including food quality, dietary diversity, cooking styles, and nutrient absorption. An imbalance in these aspects can lead to suboptimal nutrition, negatively impacting health. In recent decades, diets have shifted toward increased consumption of processed, high-calorie foods and reduced intake of fruits and vegetables, contributing to the rise of diet-related diseases \cite{bu2024recognition, he2024health}.

To counteract these trends, dietary guidelines such as the Mediterranean and Japanese diets have been widely promoted for their health benefits and balanced nutritional profiles \cite{singh2022and, zhang2022influence}. However, these guidelines provide general recommendations and require individualized monitoring to assess effectiveness. Traditional dietary assessment methods, such as 24-hour recall and food frequency questionnaires, often fail due to reliance on self-reporting, which can be tedious and prone to inaccuracies. Advancements in dietary assessment suggest that automatic and personalized nutrition may offer more effective solutions \cite{linseisen2025perspective, yang2024chatdiet}. 

\begin{figure*}
    \centering
    \includegraphics[width=0.95\linewidth]{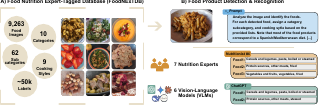}
    \caption{Overview of the proposed framework. (A) The FoodNExTDB consists of 9,263 food images labeled by nutrition experts across 10 food categories, 62 subcategories, and 9 cooking styles, with approximately 50,000 assigned labels. (B) The experimental setup, where six Vision-Language Models (VLMs) process food images using a structured prompt and generate predictions. These predictions are then compared against expert annotations to assess model performance in food product detection and recognition.}
    \label{fig:graphabstr}
\end{figure*}

Food computing has emerged as a rapidly growing field, applying computational techniques to various food-related domains \cite{foodcomputing}. The increasing digitalization of daily life has led to the availability of vast amounts of online food data, enabling the development of large-scale food datasets \cite{rodriguez2023dining}. Moreover, advancements in artificial intelligence (AI) have facilitated the evaluation of nutritional content and the understanding of individual dietary habits, contributing to both health maintenance and disease prevention \cite{sheng2024lightweight,romero2023ai4foodfw}. machine learning (ML), particularly deep learning (DL), has demonstrated remarkable capabilities in food image recognition and classification tasks \cite{luo2023research, romero2024leveraging}. However, these methods primarily focus on visual classification and struggle to extract contextual information, such as ingredient composition, preparation techniques, etc.

To address these limitations, foundation models have demonstrated impressive performance across a wide range of multimodal tasks \cite{achiam2023gpt, team2023gemini, deandres2024good}. Models such as ChatGPT and Gemini have achieved state-of-the-art results in applications such as food tracking, dietary recommendation, and food science research \cite{yang2024chatdiet, qi2023foodgpt, zhang2024llm, MA2024104488, pitsilou2024using, liu2024deepseek}. More recently, Vision-Language Models (VLMs) have emerged as the next generation of explainable computer vision systems, integrating textual and visual data for a more comprehensive understanding of food-related information \cite{lin2024vila, li2023blip, sharma2024losing}. Despite their potential, several challenges still remain such as the reliability, interpretability, and accuracy of these models for food-related tasks \cite{bordes2024introduction}. This raises fundamental questions about their practical application in real-world nutritional assessments: \textit{Are VLMs ready to assist nutrition experts in critical tasks such as supervised dietary quality assessment? Can they accurately recognize food products based on subtle factors like cooking styles, which can significantly impact nutritional values? Can current AI-based tools reliably evaluate an individual's dietary behavior solely through image uploads?} 

Advancing knowledge around these questions, this paper has the following contributions: \textit{i)} we introduce the Food Nutrition Expert-Tagged Database (FoodNExTDB), a unique food image database that contains 9,263 expert-labeled images, containing real dietary records from individuals who participated in a weight loss intervention \cite{romero2023ai4fooddb}, \textit{ii)} we propose an experimental framework to assess state-of-the-art VLMs, analyzing their capabilities in food recognition at different levels and prompt comprehension, and \textit{iii)} we design a novel evaluation metric, Expert-Weighted Recall (EWR), that accounts for inter-annotator variability. Differences in nutritional paradigms, dietary guidelines, and individual professional experiences can lead to inconsistencies in labeling the same food item. Also, cultural and regional differences further contribute to variations in annotation. The FoodNExTDB includes annotations from seven different nutrition experts (50,000 labels), representing a valuable contribution for the research community. Figure \ref{fig:graphabstr} illustrates the overall framework of the present study.

The remainder of this paper is structured as follows: Sec. \ref{sec:rel} reviews key food image databases and recognition systems. Sec. \ref{sec:db} details the proposed FoodNExTDB. Sec. \ref{sec:methods} presents the VLM models used in this study, while Sec. \ref{sec:experimental_framework} outlines the experimental protocol and proposed metric. Sec. \ref{sec:results} analyzes the experimental results, followed by discussions and conclusions in Sec. \ref{sec:discussion} and \ref{sec:conclusion}, respectively.

\section{Related Works}
\label{sec:rel}

\subsection{Food Image Databases}

Food image databases are essential in training and evaluating automatic recognition models. These databases are typically acquired through three main methods:  
\textit{i)} \textit{self-collected}, where food images are manually captured in controlled or semi-controlled environments,  \textit{ii)} \textit{web-scraped}, where large-scale databases are gathered from online platforms, and  \textit{iii)} \textit{combination}, where multiple existing databases are merged to enhance diversity and coverage.  

First, self-collected databases provide high-quality images but are limited in size due to manual acquisition. For instance, the UNIMIB2016  database includes around 20K images of various food products \cite{unimib2016}, while Nutrition5k  offers 5K unique dishes with videos and nutritional data \cite{thames2021nutrition5k}. 

%Similarly, UNICT-FD1200 \cite{unict} features food images representing diverse cuisines, covering both Western and Eastern dishes.

In contrast, web scraping allows for the collection of large-scale food image databases from social media and online sources. Examples include Food-101 \cite{Food101} with 101,000 images spanning 101 food categories, VireFood-172, with 172 categories \cite{chen2016deep},  and ISIA Food-500 \cite{ISIA}, which covers 500 different food types. 

%Other large-scale databases include VIPER-FoodNet \cite{viper} with 18,000 images and Recipe1M+ \cite{marin2021recipe1m}, a multimodal dataset containing over 13 million images and 1 million recipes. The Food2K \cite{food2k} dataset further expands this collection with 1 million images categorized into 2,000 food types.  

Combination databases merge multiple sources to create more diverse and comprehensive databases. For example, Food-11 \cite{EPFL} integrates data from Food-101, UECFOOD-100 \cite{UECFood100}, and UECFOOD-256 \cite{UECFood256}, grouping food items into 11 main categories. MAFood-121 \cite{mafood} focuses on global cuisines, featuring 121 food products across 21,000 images. AI4FoodNutritionDB  introduces a structured nutritional taxonomy, consisting of 553,000 images categorized into different nutritional levels, main categories, subcategories, and final food products \cite{romero2024leveraging}. Finally, Food-500 Cap enhances traditional food databases by incorporating 24,700 images with detailed captions that describe fine-grained visual attributes \cite{ma2023food}.

\subsection{Food Image Recognition Models}

Food image recognition has evolved significantly, transitioning from traditional classification techniques to DL approaches. Early methods struggled with complex and diverse food datasets, leading to the adoption of DL architectures such as convolutional neural networks (CNNs). These models integrated feature extraction and classification, achieving high accuracy rates exceeding 80\% \cite{food2k, results5k}. Outstanding models include Squeeze-and-Excitation Networks (SENet) \cite{senet}, Stacked Global-Local Attention Networks (SGLANet), and Progressive Region Enhancement Networks (PRENet), which demonstrated strong generalization across multiple food databases \cite{food2k, ISIA}. Transfer learning and ensemble techniques further enhanced food classification performance, while object detection frameworks such as LOng-tailed FIne-Grained Network (LOFI) and YOLO improved food recognition \cite{rodriguez2024lofi, bu2024recognition, luo2023research}.

The introduction of vision transformers (ViTs) marked a shift in food recognition by capturing global dependencies within images, enhancing classification accuracy \cite{sheng2024lightweight}. However, ViTs alone required significant computational resources and struggled with fine-grained classification due to high intra-class variations in food appearance. As a result, hybrid architectures combining CNNs and ViTs emerged, leveraging the spatial awareness of CNNs and the contextual understanding of transformers \cite{nfor2025explainable}. 

In the present study, we explore the application of recent VLMs. Models such as CLIP align images and text in a shared multimodal space, enabling zero-shot classification and improved fine-grained food differentiation \cite{radford2021learning, ma2024integrating}. Large multimodal models like FoodLLM and Large Language and Vision Assistants (LLaVA), specifically LLaVA-Chef, extend this capability by incorporating domain-specific knowledge about food ingredients, preparation methods, and cultural contexts. These models leverage multimodal prompting, combining textual and visual inputs to enhance classification accuracy, making them effective at extracting contextual information  \cite{electronics13224552, mohbat2024llava, yin2023foodlmm}. 

%The evolution from CNNs to ViTs, hybrid models, and VLMs illustrates the increasing sophistication of food recognition systems. While CNNs and hybrid architectures remain valuable for structured image classification, VLMs provide a more flexible and semantically enriched approach, capable of understanding complex food compositions and culinary concepts \cite{ponte4984843multi, ma2023food}. As research progresses, integrating domain-specific ontologies and refining multimodal learning will be key to further improving food image classification performance.

\section{FoodNExTDB}\label{sec:db}

The FoodNExTDB is a food image database derived from  from the AI4FoodDB, a comprehensive multimodal database acquired from a one-month randomized controlled trial (RCT) with 100 overweight and obese participants undergoing a nutritional intervention \cite{romero2023ai4fooddb}. Figure \ref{fig:graphabstr}A) summarizes its key features, including food images collected over 14 days per participant. With many food products reflecting characteristics of Spanish cuisine, this database provides a valuable resource for studying food intake within a dietary pattern. The database is publicly available on GitHub\footnote{\href{https://github.com/AI4Food/FoodNExtDB}{https://github.com/AI4Food/FoodNExtDB}}.

%AI4FoodDB consists of multiple datasets integrating biological samples, continuous digital measurements, and dietary records collected through manual, clinical, and digital acquisition methods. For a detailed description of AI4FoodDB and the RCT, refer to \cite{romero2023ai4fooddb}

\subsection{Database Construction}

Participants were instructed to capture images of all consumed foods and beverages using their smartphones. A total of 10,739 images were collected, of which approximately 14\% were discarded during post-processing (e.g., due to non-food images, blurred images, etc.), resulting in a final database of 9,263 food images. Notably, around 88\% of the images contain a valid timestamp, a key parameter for analyzing participants' eating behaviors.

% The labeling process was done in several phases. First, 4 out of the 7 experts labeled a subset of images in order to test usability of the application and possible problems that could arise during the process. Then, two meetings were held: first, a meeting involving the full team of experts together with the application developers; second, a meeting where the experts discussed possible discrepancies that could arise during the process, with the aim of improving consistency of labels among experts. In the second phase, labelers worked with the food image dataset independently, although they could communicate with each other if necessary.

Food images were annotated during the post-processing stage by a team of seven nutrition experts, ensuring that each image was reviewed by at least three annotators. This approach was implemented to enhance labeling reliability, given the varying complexity of food product recognition. While some food products are straightforward to classify, others require expert judgment due to their ambiguity.

The annotation process involved identifying the food items present in each image and categorizing them according to a predefined nutritional taxonomy. Each food product was assigned a \textit{category}, \textit{subcategory}, and \textit{cooking style}. The taxonomy comprises 10 main food categories (e.g., \textit{``cereals and legumes", ``protein sources"}, etc.), 62 subcategories (e.g., \textit{``alcoholic beverages", ``fruits"}, etc.), and 9 cooking styles (e.g., \textit{``fried", ``boiled or stewed"}, etc.). Additionally, free-text fields were provided to accommodate food products that did not fit into the predefined classifications. The complete list of \textit{category}, \textit{subcategory}, and \textit{cooking style} classes is available in the supplementary material.

A custom Windows GUI was developed to streamline labeling while ensuring consistency. The database uniquely combines food categorization with culinary techniques, offering technical annotations and nutritional insights.

% Besides the digital DS3 - Nutrition, BGL from participants were recorded during the digital acquisition, for approximately 14 days. Specifically, they wore a Freestyle Libre 2 continuous glucose monitor (CGM) to measure BGL in mg/dL at 15-minute intervals. From $> 29,000$ hours of data, around 10\% of them were lost due to a lack of synchronization.

\subsection{Database Statistics and Characteristics}

On average, each participant captured approximately 96 food images, with a standard deviation of 58 images. Notably, $\sim$20\% of participants took fewer than 50 images, while $\sim$15\% captured more than 150. Regarding temporal distribution, most images ($\sim$79\%) were taken during Spain’s main daily meals, 1,836 images ($\sim$20\%) at breakfast, 2,540 ($\sim$27\%) at lunch, and 2,988 ($\sim$32\%) at dinner.

In total, nutrition experts assigned over 50,000 labels. The three most frequently assigned food categories are \textit{``vegetables and fruits"} ($\sim$28\%), \textit{``cereals and legumes"} ($\sim$17\%), and \textit{``beverages"} ($\sim$16\%). At the subcategory level, they are \textit{``vegetables"} ($\sim$13\%), \textit{``fruits"} ($\sim$13\%), and \textit{``bread"} ($\sim$8\%).  Finally, for cooking styles they are \textit{``none"} ($\sim$31\%), \textit{``fresh"} ($\sim$28\%), and \textit{``boiled or steamed"} ($\sim$10\%), remarking the prevalence of raw foods.

\subsection{Problems Encountered During Labeling}

The analysis of the annotations from different nutrition experts revealed inconsistencies in the classification of certain food products and cooking styles, highlighting a lack of consensus in specific cases. This variability can significantly impact dietary assessments, as cooking styles influence the nutritional quality of a meal.  

Additionally, in many cases, the cooking style could not be reliably determined from the image alone, nor could certain food components like sauces or oils, which are often visually indistinguishable. Moreover, complex dishes with multiple food categories and subcategories posed a significant challenge, making labeling tedious and requiring expert judgment for consistency.

\section{Proposed Methods}\label{sec:methods}

\subsection{Selected Vision-Language Models}\label{sec:selected_VLMs}

In this study, six state-of-the-art VLMs are explored: ChatGPT (GPT-4o), Gemini 2.0 Flash, Claude 3.5 Sonnet, Moondream, DeepSeek Janus-Pro, and LlaVA. The primary objective is to compare VLMs that are accessible through API on limited-resource environments.

ChatGPT, Gemini, and Claude are closed-source models with dedicated APIs, while DeepSeek, LlaVA, and Moondream are open-source models, which in this study are deployed using Replicate\footnote{\href{https://replicate.com/}{https://replicate.com/}}, a platform offering efficient AI infrastructure to open-source VLMs through an API. To ensure unbiased results, each model instance is initialized independently for each image.

Several parameters can be adjusted by the user in models that support customization, including \textit{temperature}, \textit{top-p}, \textit{max tokens}, and \textit{seed}. The \textit{temperature} parameter controls the randomness of model outputs, where lower values result in more deterministic responses. The \textit{top-p} parameter, also known as nucleus sampling, defines the probability mass for selecting the next token, balancing diversity and coherence. \textit{Max tokens} determines the maximum length of the generated output, while \textit{seed} ensures reproducibility across different runs. For models that allow parameter tuning, we experimentally set the following values: \textit{temperature} = 0.2, \textit{top-p} = 0.95, \textit{max tokens} = 200, and \textit{seed} = 42. 

We describe next the key details of the selected VLMs:

\begin{itemize}
    \item \textbf{ChatGPT (GPT-4o)}: this is a multimodal model from OpenAI, capable of processing both text and image inputs while generating text-based outputs. It has been pre-trained on a diverse dataset, including web content, proprietary data, and multimodal sources \cite{hurst2024gpt}. For this study, GPT-4o is accessed via OpenAI’s API using the \texttt{openai} Python library. It supports a 128,000-token context window and a maximum output of 16,384 tokens\footnote{\href{https://chatgpt.com/}{https://chatgpt.com/}}.
    \item \textbf{Gemini 2.0 Flash}: developed by Google, this is a high-speed multimodal model designed for diverse tasks, including text and image processing. It features next-generation capabilities such as native tool use and multimodal generation. The model has been pre-trained on a diverse dataset, enabling strong vision-language reasoning \cite{team2023gemini, team2024gemini}. For this study, Gemini is accessed via Google’s API using the \texttt{google} Python library. It supports a context window of around 1M tokens and a maximum output of 8,192 tokens\footnote{\href{https://deepmind.google/}{https://deepmind.google/}}.
    \item \textbf{Claude 3.5 Sonnet}: developed by Anthropic, this is a state-of-the-art model with multilingual, vision, and reasoning capabilities. It features a 200,000-token context window and a maximum output of 8,192 tokens \cite{anthropic2024claude}. Trained on a proprietary dataset combining publicly available web data, third-party sources, and internally generated content, Claude employs a hybrid reasoning approach with strong performance in code generation, computational tasks, and extended context processing. For this study, it is accessed via Anthropic’s API\footnote{\href{https://claude.ai/}{https://claude.ai/}}.
    \item \textbf{Moondream}: this is an open-source, lightweight VLM designed for efficient image analysis, object detection, visual reasoning, and scene comprehension. Moondream-2B, in particular, is optimized for visual understanding tasks with a minimal computational footprint\footnote{\href{https://moondream.ai/}{https://moondream.ai/}}. For this study, we use Moondream-2B via Replicate’s API\footnote{\href{https://replicate.com/lucataco/moondream2}{https://replicate.com/lucataco/moondream2}}.
    \item \textbf{DeepSeek Janus-Pro}: this is one of the first open-source models incorporating a vision module. It is built upon DeepSeek Janus \cite{wu2024janus}, an autoregressive transformer framework designed for both multimodal understanding and generation. The core innovation lies in decoupling visual encoding, enhancing the model’s ability to process and generate text from visual inputs\footnote{\href{https://chat.deepseek.com/}{https://chat.deepseek.com/}} \cite{chen2025janus}. For this study, DeepSeek Janus-Pro is accessed via Replicate’s API\footnote{\href{https://replicate.com/deepseek-ai/janus-pro-1b}{https://replicate.com/deepseek-ai/janus-pro-1b}}.
    \item \textbf{LlaVA}: this is an end-to-end trained multimodal model that employs a fully-connected vision-language connector\footnote{\href{https://LlaVA-vl.github.io/}{https://LlaVA-vl.github.io/}} \cite{liu2024improved}. For this study, we use the \texttt{LlaVA-v1.6-mistral-7b} model via Replicate’s API\footnote{\href{https://replicate.com/yorickvp/LlaVA-v1.6-mistral-7b}{https://replicate.com/yorickvp/LlaVA-v1.6-mistral-7b}}. Leveraging Mistral-7B, LlaVA enhances multimodal text generation and image-based reasoning while balancing performance and computational cost.

\end{itemize}

\subsection{Prompt Design}\label{sec:prompt_design}

To optimize the prompt for obtaining accurate and structured responses, several key aspects were considered. The primary goal was explicitly defined: analyzing the image and identifying the foods present. Since many of the food products align with Mediterranean and Spanish dietary patterns, the final prompt enforces a standardized format, requiring each detected food item to be categorized into a predefined \textit{category}, \textit{subcategory}, and \textit{cooking style}. To maintain consistency, it restricts responses to the given taxonomy, prevents additional explanations or assumptions, and ensures outputs follow a strict, structured format. 

Some \textit{categories} are \textit{``cereals and legumes", ``vegetables and fruits", ``protein sources"}, etc. Furthermore, \textit{subcategories} such as \textit{``fruits", ``infusions", and ``pizza"}, and \textit{cooking styles}, including \textit{``fresh", ``fried", ``boiled" or ``steamed", and ``oven-baked"}, ensure a detailed and standardized representation of food items.

To ensure consistency, foods not in the predefined list were labeled as \textit{``Others"}, while partially visible or distant items were excluded. Finally, for models that failed to generate structured responses (i.e., DeepSeek, LLaVA, and Moondream), we implemented a secondary post-processing step using ChatGPT-4o. This step refined outputs by correcting formatting errors, separating merged elements, and classifying food products into predefined \textit{category}, \textit{subcategory}, and \textit{cooking style}. The complete prompts are available in the supplementary material.

\section{Experimental Framework}\label{sec:experimental_framework}

\subsection{Experimental Protocol}\label{sec:protocol}

In order to systematically evaluate the performance of the six selected VLMs with our proposed FoodNExTDB, we design an experimental protocol consisting of three tasks. As indicated in Sec. \ref{sec:prompt_design}, we insert to each VLM a structured prompt to analyze each image, identifying individual food products, and generating an appropriately formatted output. To ensure consistency, manual post-processing was applied to fewer than 1\% of the generated labels, addressing cases where models produced similar but non-exact matches (e.g., generating food categories or cooking styles not included in the predefined taxonomy). The total cost of all experiments was \$434.04, covering all API executions. We describe next the three tasks analyzed in our experimental study.

\subsubsection{Task 1: Food Image Recognition}

This task evaluates the ability of VLMs to recognize food products from images by classifying them into predefined categories. The evaluation is conducted at three consecutive levels, from simpler to more complex classifications:  

\begin{itemize}
    \item \textit{Category}: Recognizing general food groups, such as \textit{``vegetables and fruits"} or \textit{``protein sources"}.
    \item \textit{Category + Subcategory}: Providing finer-grained classification within each category, for example, distinguishing between \textit{``vegetables"} and \textit{``fruits"} subcategories.
    \item \textit{Category + Subcategory + Cooking Style}: Identifying both the food \textit{category} and \textit{subcategory} and its preparation method, such as differentiating between \textit{``grilled"} fish and \textit{``fried"} fish.
\end{itemize}

\begin{figure*}
    \centering
    \includegraphics[width=0.95\linewidth]{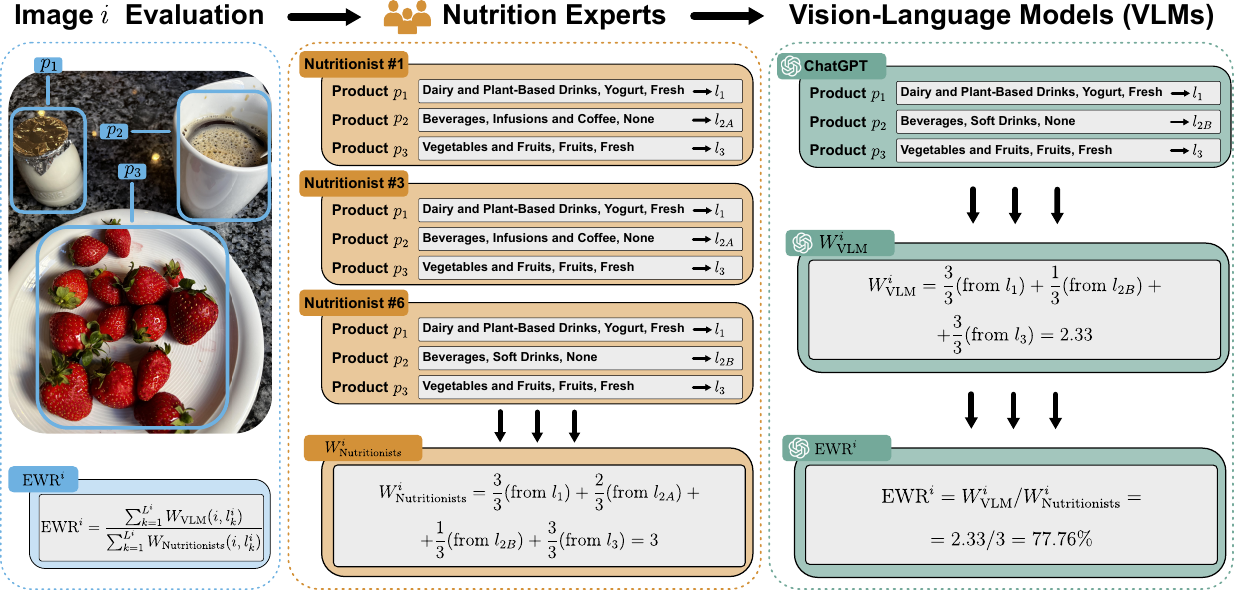}
    \caption{Illustration of the proposed Expert-Weighted Recall (EWR) computation for a food image \( i \) (left). This graphical example compares annotations (e.g. label $l_1=\textit{Yogurt}$ for product $p_1$) from three nutrition experts (middle) with the predictions made by a VLM (right). The proposed EWR metric reflects how well the VLM aligns with expert consensus while accounting for annotation variability.}
    \label{fig:fig2b}
\end{figure*}

\subsubsection{Task 2: Fine-Grained Food Recognition}
The second task focuses on the model's ability to recognize specific food products correctly. Performance is analyzed across different categorization levels to determine which food products are easier or more challenging to recognize. This task helps to assess whether VLMs can reliably differentiate between foods that look similar and correctly assign them to their respective \textit{categories} and \textit{subcategories}.

\subsubsection{Task 3: Image Complexity Performance}
The third task examines the impact of image complexity on model performance. Two scenarios are evaluated, with images containing only one identifiable food item (\textit{single-product} images), and images featuring multiple food products in a single image (\textit{multi-product} images).

\subsection{Proposed Evaluation Metric: EWR}

Our metric quantifies how well the VLMs match the annotations considering possible differences between annotators. In image $i$,  the annotators identify $p_{j}^i$ food products, with $j=1,\ldots,M^i$, where $M^i$ is the total number of food products detected by  the annotators. Let $L^i$ be the total number of product labels $l_k^i$, with $k=1,\ldots,L^i$, assigned by the annotators (note that a single product can originate multiple labels from multiple experts). We assign a different weight for each marked label $l_k^i$ based on the level of label agreement between nutrition experts $ W_{\text{Nutritionists}}(i,l_k^i)=n_k^i/N^i$, where $n_k^i$ is the number of nutritionists who marked the given label $l_k^i$ for the same product $p_j^i$ (out of a total of $N^i$ annotators). Note that the number of products and annotators can vary between images, and some annotators may not label a particular product if they are not sure about it.

For VLMs predictions we consider the following weights: \textit{i)} if a predicted product label appears in the annotations of experts, it is assigned the corresponding weight (i.e., $W_{\text{VLM}}(i,l_k^i)=n_k^i/N^i$), otherwise \textit{ ii)} if a prediction is not in the annotated list, then $W_{\text{VLM}}(i,l_k^i)=0$. The propose Expert-Weighted Recall (EWR) metric is as follows:

\begin{equation} \text{\text{EWR}}^i =  \frac{ \sum_{k=1}^{L^i} W_{\text{VLM}}(i,l_k^i)}{ \sum_{k=1}^{L^i} W_{\text{Nutritionists}}(i,l_k^i)} \end{equation}

\begin{table}[t]
\centering

\resizebox{0.95\columnwidth}{!}{%
\begin{tabular}{llcccc}
\hline
\multicolumn{2}{c}{\textbf{\small{Model Name}}} & \textbf{\small{Category}} & \textbf{\small{+Subcat.}} & \textbf{\begin{tabular}[c]{@{}c@{}}\small{+Cooking}\\ \small{Style}\end{tabular}} & \textbf{\small{Average}} \\ \hline
\multirow{3}{*}{\begin{tabular}[c]{@{}l@{}}\rotatebox[origin=c]{90}{\small{\textbf{Closed}}} \rotatebox[origin=c]{90}{\small{\textbf{Source}}}\end{tabular}} & ChatGPT & 80.67 & 69.87 & 42.41 & 64.32 \\
 & Gemini & \textbf{85.79} & \textbf{74.69} & \textbf{50.00} & \textbf{70.16} \\
 & Claude & 82.60 & 69.88 & 45.09 & 65.86 \\\hdashline
\multirow{3}{*}{\begin{tabular}[c]{@{}l@{}}\rotatebox[origin=c]{90}{\small{\textbf{Open}}} \rotatebox[origin=c]{90}{\small{\textbf{Source}}}\end{tabular}} & Moondream & 76.94 & 63.28 & 23.91 & 54.71 \\
 & DeepSeek & 49.39 & 37.42 & 15.30 & 34.04 \\
 & LlaVA & 63.67 & 48.84 & 28.48 & 47.00 \\ \hline
\end{tabular}%
}

\caption{Performance comparison in terms of Expert-Weighted Recall (EWR) of the selected VLMs in food image recognition. Results are reported across three categorization levels, from simpler to more complex ones: \textit{i)} \textit{Category}, \textit{ii)} \textit{Category + Subcategory}, and \textit{iii)} \textit{Category + Subcategory + Cooking Style}.}
\label{tab:results}
\end{table}

The final EWR score is obtained by averaging the individual $\text{EWR}^i$ values from all images. The EWR metric ensures that higher-agreement predictions contribute more to the final score, allowing flexibility for partial agreement. Fig.~\ref{fig:fig2b} shows the EWR computation by comparing the annotations (middle) with the VLM predictions (right). In the food image \( i \) (left), three food products are detected: $p_1^i$, $p_2^i$, and $p_3^i$ (in Fig.~\ref{fig:fig2b} we remove the upper $i$ notation for simplicity). All experts agreed on \( p_1 \) (label $l_1=\textit{``yogurt"}$), while \( p_2 \) was labeled differently by two experts. All identified \( p_3 \) with the label $l_3=\textit{``fruits"}$. ChatGPT correctly predicted \( l_1 \), matched \( l_{2B} \), and identified \( l_3 \). The 77.76\% EWR achieved reflects how well ChatGPT aligns with the experts while accounting for annotation variability.

\section{Results}\label{sec:results}

This section evaluates the VLM performance in food image recognition. Sec. \ref{sec:task1} analyzes classification across different granularity levels, Sec. \ref{sec:task2} explores the most and least challenging food products, and Sec. \ref{sec:task3} evaluates the impact of image complexity on recognition accuracy.

\begin{figure*}
    \centering
    \includegraphics[width=0.9\linewidth]{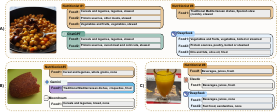}
    \caption{Examples of VLMs predictions compared to nutritionist's annotations. (A) A multi-component dish where some experts identify individual ingredients (\textit{``legumes"}, \textit{``meat"}, \textit{``vegetables"}), while others classify it as \textit{``Spanish stew (\textit{cocido}"}) (B) A whole grain bread misclassified by Gemini as \textit{``croquettes"} due to shape similarity. (C) An orange juice image where some VLMs such as DeepSeek incorrectly identifies a sandwich from a background image on the paper tray liner.}

    \label{fig:fig2}
\end{figure*}

\subsection{Task 1: Food Image Recognition}\label{sec:task1}
Table \ref{tab:results} presents the performance of each VLM in terms of EWR across three classification levels: \textit{i) category}, \textit{ii) category + subcategory}, and \textit{iii) category + subcategory + cooking style}. Overall, closed-source models (ChatGPT, Gemini, and Claude) consistently outperform open-source ones (Moondream, DeepSeek, and LLaVA) at all classification levels. Gemini achieves the highest average EWR across all levels (70.16\%), indicating a broader detection range, while ChatGPT (64.32\%) and Claude (65.86\%) maintain strong performance.

As classification complexity increases, all models experience performance drops. For instance, Gemini’s EWR declines from 85.79\% (\textit{category}) to 74.69\% (\textit{category + subcategory}) and 50.00\% (\textit{category + subcategory + cooking style}), with a similar trend observed for ChatGPT (80.67\%, 69.87\% and 42.41\%, respectively). These results highlight the challenge of fine-grained recognition.

Among open-source models, Moondream outperforms DeepSeek and LLaVA (avg. 54.71\% vs. 34.04\% and 47.00\%, respectively), excelling in \textit{category} and \textit{category + subcategory} recognition. DeepSeek, with the lowest average EWR (34.04\%), struggles due to limited exposure to food datasets. Figure \ref{fig:fig2} presents challenging cases. In Figure \ref{fig:fig2}A), a multi-component dish is inconsistently labeled by experts, with VLMs only recognizing individual ingredients. Figure \ref{fig:fig2}B) shows whole-grain bread misclassified by Gemini as croquettes due to shape similarity. In Figure \ref{fig:fig2}C), DeepSeek mistakenly identifies a sandwich from a background image on a paper tray liner, highlighting VLM susceptibility to background distractions.

% Discussion
% The results highlight the effectiveness of proprietary models in structured food recognition tasks while emphasizing the challenges faced by open-source models in comprehending complex prompts and domain-specific categorizations.

\subsection{Task 2: Fine-Grained Food Recognition}\label{sec:task2}

\begin{figure*}[t]
    \centering
    \includegraphics[width=0.95\linewidth]{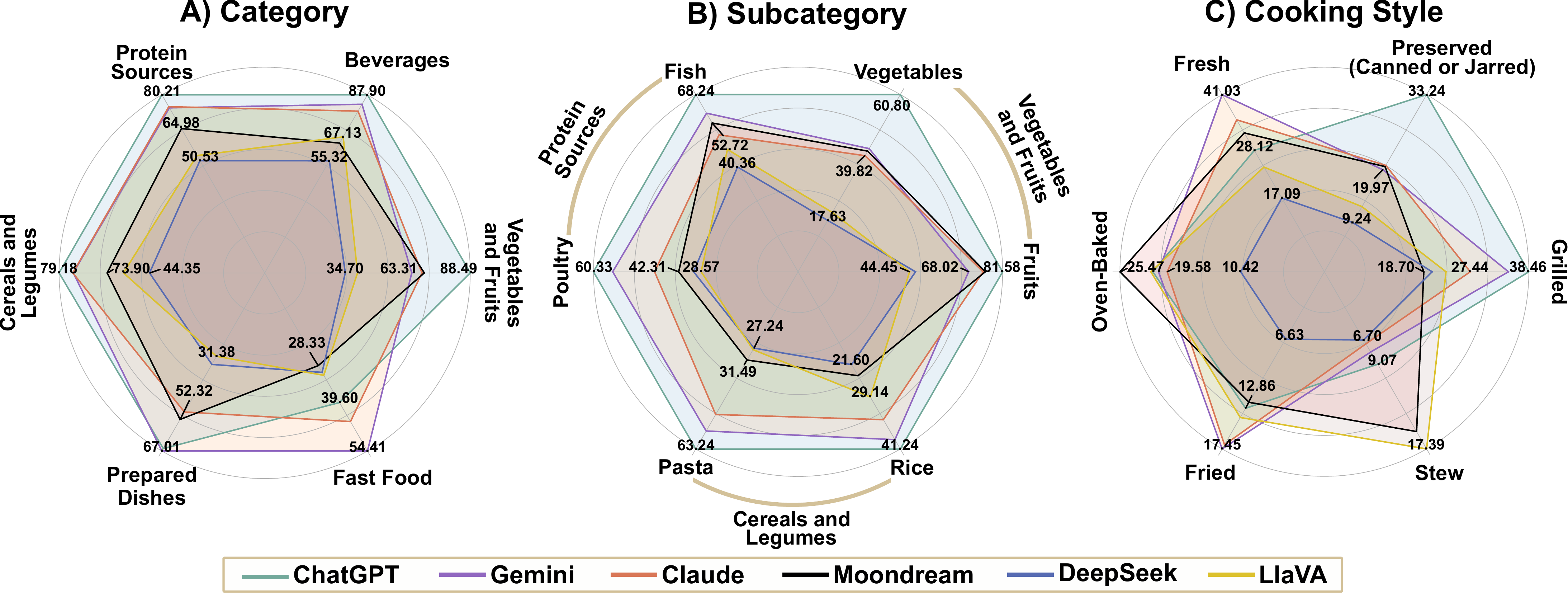}
   \caption{Radar charts illustrating VLM performance in fine-grained food recognition. We include some examples of all available classes considered in our proposed FoodNExTDB, i.e., 10 categories, 62 subcategories, and 9 cooking styles. (This figure is best viewed in color.)}
    \label{fig:fig3}
\end{figure*}

The ability of VLMs to recognize specific food products varies significantly across different categorization levels. Figure \ref{fig:fig3} presents radar charts illustrating the percentage of correctly recognized food products for each level, similar to Task 1. It is important to highlight that these radar charts include some examples of all possible classes considered in our proposed FoodNExTDB, i.e., 10 categories, 62 subcategories, and 9 cooking styles.

At the \textit{category} level, VLMs perform best in recognizing food categories such as \textit{``cereals and legumes", ``protein sources"}, and \textit{``dairy and plant-based drinks"}. In these particular categories, ChatGPT consistently outperforms other models, achieving the highest accuracy across most categories. Conversely, \textit{``fast food"} remains the most challenging category for all models. At the \textit{subcategory} level, recognition declines, with \textit{``fruits"} being more accurately detected than \textit{``vegetables"}  and \textit{``fish"} outperforming \textit{``poultry"} within their corresponding main categories. \textit{``Pasta"} is also more frequently recognized than \textit{``rice"}. 

At the \textit{cooking style} level, all models struggle significantly. \textit{``Fresh"} is the most identifiable style, followed by \textit{``grilled"}, while \textit{``fried"} and \textit{``stewed"} are the least accurately predicted. Notably, Gemini excels in detecting \textit{``fresh"} foods, Moondream in \textit{``oven-baked"} products, ChatGPT in \textit{``preserved"} foods, and LLaVA in \textit{``stewed"} dishes, despite overall lower performance in this category.

\begin{table}[t]
\centering

\resizebox{\columnwidth}{!}{%
\begin{tabular}{clccc}
\hline
\multicolumn{2}{c}{\textbf{\small{Model Name}}} &  \textbf{\small{Single-Product Image}} & \textbf{\small{Multi-Product Image}} & \textbf{\small{Average}} \\ \hline
\multirow{3}{*}{\begin{tabular}[c]{@{}l@{}}\rotatebox[origin=c]{90}{\small{\textbf{Closed}}} \rotatebox[origin=c]{90}{\small{\textbf{Source}}}\end{tabular}} &ChatGPT & 90.71 & 76.51 & 83.61 \\
&Gemini  & \textbf{94.52} & \textbf{82.18} & \textbf{88.35}\\
&Claude & 92.48 & 78.51 & 85.50\\\hdashline
\multirow{3}{*}{\begin{tabular}[c]{@{}l@{}}\rotatebox[origin=c]{90}{\small{\textbf{Open}}} \rotatebox[origin=c]{90}{\small{\textbf{Source}}}\end{tabular}} &Moondream & 90.03 & 71.52 & 80.78\\
&DeepSeek  & 43.48 & 51.83 &47.66 \\
&LlaVA & 68.29 & 61.76 &65.03\\ \hline
\end{tabular}%
}

\caption{Performance comparison in terms of EWR of the selected VLMs in food recognition (\textit{category} level) based on the image complexity (i.e., \textit{single-product} and \textit{multi-product} images).}
\label{tab:img_complexity}
\end{table}

\subsection{Task 3: Image Complexity Performance}\label{sec:task3}

Finally, we evaluate the impact of image complexity (i.e., the number of food products present in one image) on the VLMs' performances. This analysis is carried out for \textit{single-product} images (i.e., only one food product appears in the image) and \textit{multi-product} images (i.e., multiple food products can appear in one image). Table \ref{tab:img_complexity} presents the EWR of each VLM across these scenarios. For all models except DeepSeek, performance is higher for the scenario of single-product images, with VLMs such as ChatGPT, Gemini, Claude, and Moondream are able to achieve EWRs above 90\%. In contrast, EWR results decrease considerably for the scenario of multi-product images. Again, Gemini demonstrates the highest overall performance in both scenarios (88.35\% EWR) whereas DeepSeek is the worst (47.66\% EWR), showing a difference of over 20\% EWR with the rest of the VLMs. 

A more detailed analysis of our proposed database reveals that, for the scenario of single-product images, approximately 28\% of images contain \textit{``fruits"}, 10\% include \textit{``beverages"} such as \textit{``infusions or coffee"}, and 7\% feature \textit{``yogurt and fresh cheese"}. As observed in previous experiments, these are among the most accurately recognized food \textit{categories} and \textit{subcategories}, which explains the higher performance.

\section{Discussion and Future Work}\label{sec:discussion}
Diet analysis remains a major challenge in nutrition, requiring the consideration of multiple interrelated factors. While pure image recognition models have improved significantly in the task of food recognition, they still struggle with complex, multi-food images and fail to provide sufficient contextual understanding. Although these models bring automated nutritional assessment closer to reality, they remain insufficient for a comprehensive analysis.

VLMs present a promising alternative by integrating textual and visual reasoning, improving explainability in food recognition and dietary analysis. However, they still face difficulties with fine-grained tasks such as identifying cooking style, which require additional multi-modal data integration. More research is needed to benchmark their performance against transformer-based models in this area.

A key observation is the performance gap between open- and closed-source VLMs. Open-source models consistently underperform, often struggling with structured prompts and generating accurate responses. As a result, improving fine-tuning strategies, dataset diversity, and domain-specific training are crucial to bridge this gap.

Finally, integrating VLM with personalized nutrition strategies could improve dietary tracking and chronic disease prevention. Combining AI-driven food recognition with multimodal data from wearables, dietary questionnaires, and expert supervision may improve accuracy and adherence to automated dietary assessments \cite{romero2024personalized}.

\section{Conclusion}\label{sec:conclusion}
This study presents FoodNExTDB, a food image database with 9,263 expert-labeled images, many reflecting Mediterranean and Spanish diets. A key strength of this database is that all images were annotated by seven nutrition experts, adding value by providing structured nutritional information, including the main \textit{category}, \textit{subcategory}, and \textit{cooking style} for each detected food product.

We also propose an experimental framework to assess six state-of-the-art VLMs, analyzing their capabilities in food recognition and prompt comprehension. Due to inter-annotator variability, we designed a novel weighted evaluation metric named Expert-Weighted Recall (EWR). Our findings reveal a clear gap in model performance across classification levels. While in main categories such as \textit{``protein sources"} and \textit{``vegetables and fruits"} the top models reach around 80\% EWR, \textit{cooking style} recognition remains challenging, with top VLMs achieving only 50\% EWR. This discrepancy highlights VLMs' limits in capturing fine-grained food attributes, suggesting the need for advances in context-aware learning and multimodal integration.

\section{Acknowledgments}
This study is supported by projects: AI4FOOD-CM (Y2020/TCS6654), FACINGLCOVID-CM (PD2022- 004-REACT-EU), INTER-ACTION (PID2021-126521OB-I00 MICINN/FEDER), HumanCAIC (TED2021-131787BI00 MICINN), PowerAI+ (SI4/PJI/2024-00062), and Cátedra ENIA UAM-VERIDAS en IA Responsable (NextGenerationEU PRTR TSI-100927-2023-2). We thank the nutrition experts for their valuable image annotations (B. Lacruz-Pleguezuelos, L.J. Marcos-Zambrano, S. Gallego Pozo, J. Haya, M. Izquierdo-Muñoz, J.A. Lemke Flores, G.X. Bazán).

{
    \small
    \bibliographystyle{ieeenat_fullname}
    \bibliography{main}
}


\end{document}